\pdfoutput=1
\documentclass[11pt]{article}

\usepackage{acl}

\usepackage{times}
\usepackage{latexsym}
\usepackage{amsmath}
\usepackage{multirow}
\usepackage{tabularx}
\usepackage{hyperref}

\usepackage[T1]{fontenc}
\usepackage[utf8]{inputenc}

\usepackage{microtype}

\usepackage{graphicx}

\definecolor{todo-color}{rgb}{0, 0.8, 0.6}

\title{Comparative Snippet Generation}

\author{Saurabh Jain, Yisong Miao, Min-Yen Kan \\
        National University of Singapore \\
        \\
        \texttt{saurabhjain@u.nus.edu, miaoyisong@gmail.com,}\\ \texttt{kanmy@comp.nus.edu.sg}}

\begin{document}
\maketitle
\begin{abstract}
% Min: This is introduction, not abstract.
%Online reviews have become a prominent medium for users to express their opinions regarding products or services.
% Such vast numbers of reviews are equally important for buyers, sellers, and manufacturers.
%Although e-commerce platforms provide information regarding a product or service and template-based functionality to compare products, these comparisons do not include information present in the reviews.
% Min: actually start here
% Orig: Our work attempts to bridge the gap by modelling reviews to generate comparative responses consisting of positive and negative experiences regarding the product.
We model product reviews to generate comparative responses consisting of positive and negative experiences regarding the product. 
Specifically, we generate a single-sentence, comparative response from a given positive and a negative opinion.
We contribute the first dataset for this task of Comparative Snippet Generation from contrasting opinions regarding a product, and a performance analysis of a pre-trained BERT model to generate such snippets.
% To the best of our knowledge, no previous work has been done to generate comparative responses from review opinions.
\end{abstract}

\section{Introduction}
The proliferation of opinions on the Web has transformed the way users express their opinions and experiences about aspects of products and services.
\emph{Online user reviews} contribute personal opinions, and when aggregated together, 
these reviews play a crucial role in purchasing decisions.
However, due to the large volume of reviews, it can be infeasible for customers to skim all such sources.
As users have to navigate through a large pool of opinions to make decisions, \emph{opinion mining and summarization} grows in importance.  As such, this area of work has received significant attention.

\begin{figure}
    \centering
    \includegraphics[width=\columnwidth]{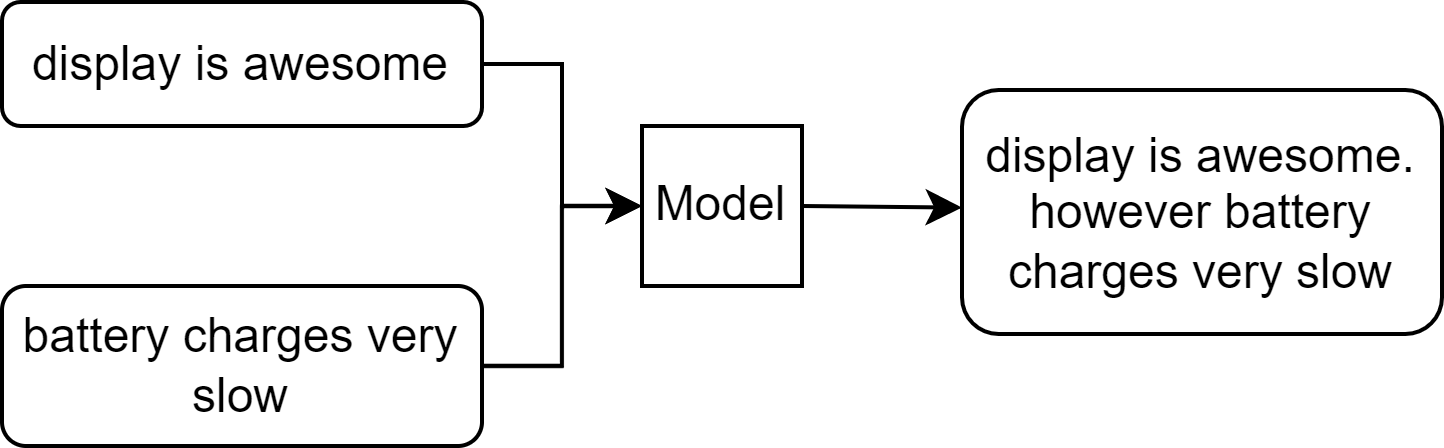}
    \caption{Comparative Snippet Generation: taking a positive and a negative opinion and generating a comparative response.}
    \label{fig:model_view}
\end{figure}

Many e-commerce platforms provide functionalities to compare products.
These functionalities may be template-based and compare products on the basis of information provided by sellers.
Comparative opinions from customers, who are experienced users of the product or service, are largely missing from such template-based comparison.
On the other hand, question answering systems based on reviews, such as AmazonQA \cite{Gupta2019AmazonQAAR}, often only tell one side of the story in the response: either positive or negative.
In our opinion, there is a demand for compact representations of both positive and negative opinions of products.  Such compact textual representation could be enunciated by dialogue agents or shown as a succinct summary to drill-down on in a mobile interface.
To the best of our knowledge, no such work has been done yet to provide \emph{comparative responses} regarding a product to a user.
We attempt this novel task.
We take as input a positive and a negative opinion regarding a product and generate a comparative, single-sentenced fused response, which we call a comparative snippet (Fig.~\ref{fig:model_view}).

We extract single-sentence summaries of positive and negative opinions, separately, from reviews of 3,269 products mentioned by the Amazon Reviews Dataset (2018).  We chose to base our corpus on this existing dataset to help spur future research on our task that can leverage existing work on the parent dataset.  
We then combine these positive and negative opinions to generate comparative responses.
Our final dataset contains \(174,394\) training instances, \(19,725\) validation instances, and \(21,397\) test instances\footnote{\url{https://github.com/WING-NUS/comparative-snippet-generation-dataset}}.
% Min: changed footnote
% \footnote{\url{https://github.com/saurabhhssaurabh/product-comparison-dataset}}.
% \footnote{\url{https://github.com/saurabhhssaurabh/comparative-snippet-generation-dataset}}.
We also successfully model positive and negative opinions to generate a comparative response expressing both positive and negative opinions about a target product.

\section{Related Work}
\textbf{Sentence Sentiment Detection.}
Sentiment detection classifies the opinion of a sentence into two classes, \emph{Positive} and \emph{Negative}.
Sometimes a third class, \emph{Neutral}, is also included.
Early works focus on unsupervised approaches and the use of sentiment lexicons to compute the overall sentiment of a text; e.g.,  \cite{turney-2002-thumbs}.
Subsequently, the convolutional neural network (CNN) architecture was introduced to classify the sentiment of sentences \citet{Kim2014ConvolutionalNN}. %Min: BUG add cite. %Saurabh: Done.
\citet{socher-etal-2011-semi} use recursive neural networks to learn sentiment at varying granularities (i.e., words, phrases, and sentences).
Many current well-performing neural models use the attention mechanism \cite{Vaswani2017AttentionIA, devlin-etal-2019-bert} to encode a text into a vector representation.

\textbf{Opinion Summarization.}
Opinion summarization differs from other summarization tasks in two aspects.
First, it cannot rely on reference summaries for training, as it is infeasible to get such meta-reviews.
To produce a reference summary for a single product, a reviewer may have to go through hundreds of reviews.
Second, due to the subjectivity and conflicting nature of reviews, the notion of information importance applies differently.
In this task, output summaries are based on the popularity of opinions.
Moreover to be viable, approaches must be flexible with respect to input size as products can be reviewed frequently, resulting in increasing amounts of review content.

Opinion summarization can be either \emph{abstractive} or \emph{extractive}.
In abstractive summarization, summaries are generated token-by-token to generate new sentences that articulate prevalent opinions from the inputs.  These generated summaries offer a solution to the lack of reference summaries, and can be written in the style of the input reviews.
% Min: need to cite which prior work you are talking about.
% Saurabh: Done.
However, prior work have used unrealistically small number of input reviews --- 10 or fewer --- to generate output summaries \citet{suhara-etal-2020-opiniondigest, Amplayo2021InformativeAC}.
Due to these shortcomings, we chose the alternative style of extractive summarization, which generates summaries by selecting phrases from the inputs.
As a foundation, we base our method on \citet{angelidis-etal-2021-extractive}, who used the Vector-Quantized Variational Autoencoder
(VQ--VAE) in their extractive opinion summarization.
First introduced by \citet{Oord2017NeuralDR}, VQ--VAE is used to learn discrete latent variables.
% This algorithm overcomes problems of posterior collapse and large variance associated with Variational Autoencoders.
It passes encoder output through a discretization bottleneck by lookup in the space of latent code embeddings.
Specifically, we use the Quantized Transformer ({\it cf.} \S~\ref{subsec:seg_summ}), an unsupervised neural model inspired by VQ--VAE, to generate popularity-driven opinions.
This method does not depend on vector averaging, nor does it suffer from information loss, which motivates us to use it as it easily accommodates large numbers of reviews.

\textbf{Sentence Fusion.}
Sentence fusion combines multiple sentences, which may contain redundancies, into one coherent sentence.
The output sentence not only should preserve input information but also any semantic relationships among sentences.
Sentence fusion requires understanding the discourse semantics between the input sentences.
Previously, feature-based approaches were used to combine sentences due to the lack of annotated data.
Recently, a large-scale sentence fusion dataset, DiscoFuse \citet{geva-etal-2019-discofuse}, was introduced, which has enabled the training of neural network-based models for the fusion task.
The authors also train the sequence-to-sequence model to fuse the input sentences and find that the trained model succeeds in combining the sentences through structural constructions, but performs badly when fusion involves inserting discourse connectives.
Recently, \citet{rothe-etal-2020-leveraging} uses a BERT-based encoder--decoder model.
Although this work improves the accuracy, it struggles in detecting the semantic relationships correctly between the input sentences.

Predicting discourse markers or connecting strings is a sister task of sentence fusion.
It is typically utilized as an intermediate step to improve downstream tasks.
\citet{ben-david-etal-2020-semantically} train a model to learn both the discourse relation and discourse connective together in a multi-task framework.
In our work, similar to \citet{rothe-etal-2020-leveraging}, we fuse two sentences together by training a model to learn the appropriate insertion of a discourse connective.
\section{Dataset Generation}\label{sec:dataset_generation}
An instance of our dataset contains positive and negative opinions as an input and a comparative response as an output as shown in Table \ref{tab:dataset_instance}.
Since no such dataset is available for reviews, we generate it from scratch.
Here, dataset generation includes the tasks of opinion extraction and rule-based response generation sub-tasks.
The task of opinion extraction itself includes subtasks of extraction, polarity classification, and summarization of segments.

\subsection{Opinion Extraction}
\noindent \textbf{Segment Extraction.}
A sentence of a review may contain more than one opinion.
For e.g., \textit{``display was quite bland, didn't enjoy much, but speed was brilliant.''}
This sentence contains a positive opinion, \textit{``but speed was brilliant''}, and a negative opinion, \textit{``display was quite bland''}.
Therefore, as suggested by \citet{angelidis-lapata-2018-multiple}, it is beneficial to process phrases and discourse units extracted from review sentences compared to processing these sentences directly.
Hereafter, we refer to these phrases and units as \emph{segments}.
We use work done by \citet{feng-hirst-2014-linear} to extract segments from reviews (as shown in Table~\ref{tab:edus_example}'s example).  After extracting segments, we perform the following five post-processing steps to improve overall quality:

\begin{table}[t]
    \begin{tabularx}{\columnwidth}{|c|X|}
    \hline
    \centering{Review}  &   In the end, take this tablet for what it is, a low end budget tablet that runs
Lollipop smoothly but has a less than desirable screen resolution.\\
    \hline
    \centering{\multirow{3}{*}{EDUs}}    &   In the end, take this tablet\\
    \cline{2-2}
        &   for what it is, a low end budget table\\
    \cline{2-2}
        &   that runs Lollipop smoothly\\
    \cline{2-2}
        &   but has a less than desirable screen resolution.\\
        \hline
    \end{tabularx}
    \caption{A review and its extracted segments.}
    \label{tab:edus_example}
    \bigskip
    \begin{tabularx}{\columnwidth}{|c|X|}
    \hline
    Input   &   the display is awesome. camera is not good.\\
    \hline
    Output  &   the display is awesome. however, camera is not good.\\
    \hline
    \end{tabularx}
    \caption{An instance of our dataset.}
    \label{tab:dataset_instance}
\end{table}

\begin{enumerate}
    \item We remove segments having less than three words, e.g. \textit{``good product''}, \textit{``best product''}, \textit{``very sad''}, etc.
    Such short segments are not relevant to our work.
    \item We remove leading and trailing punctuations; e.g., ``.'', ``!''. ``,'' and ``-''.
    \item We remove segments that do not contain at least one noun or pronoun and one main or auxiliary verb; e.g., \textit{``the only problem''}, \textit{``and was destroyed''} and \textit{``which is annoying''}.
    We use Spacy\footnote{\url{https://spacy.io/}} to extract a noun and a verb from a segment.
    \item Since we focus on working with segments with third-person narrative, we discard segments containing first-person words: ``i'', ``me'', ``my'', ``myself'', ``mine'', ``we'', ``us'', ``our'', ``ourselves''.  While our \emph{extractive summarization} approach (\S\ref{subsec:seg_summ}) will eventually rank such segments low, we prefer to drop these here for efficiency.
    \item If a segment starts with 
    We delete any leading occurrences of ``because'', ``and'', ``before'', ``but'', ``however'', ``now'', ``of'', ``then'', ``\&'', ``or'' from the segment.
    As an example, we edit \textit{``but it is not that great''} into \textit{``it is not that great''}, by omitting the leading \textit{``but''}.
\end{enumerate}
%
% After segment extraction, we classify each segment into a positive or negative class depending on its polarity.\\
\noindent\textbf{Segment Sentiment Classification.}
We next classify each segment into one of two categories, positive or negative.
Reviews from different domains may differ in syntactic properties --- e.g., length and vocabulary --- however, the underlying semantics and discourse properties remain the same. To the best of our knowledge, there are no segment-level polarity-annotated datasets that build from the Amazon Reviews.  As such, we use SPOT: Sentiment Polarity Annotation Dataset\footnote{\url{https://github.com/EdinburghNLP/spot-data}}, which contains \(197\) reviews taken from the Yelp \citet{tang-etal-2015-document} and IMDB \citet{Diao2014JointlyMA} datasets, annotated with segment-level polarities for positive, neutral, or negative sentiments.
AS our work only utilizes positive and negative opinions to generate a comparative response, we discard the neutral segments.
We fine-tune BERT \citet{devlin-etal-2019-bert} for polarity classification using the SPOT dataset.
Then we classify extracted segments into positive or negative class using the fine-tuned model.
\\\\
\noindent \textbf{Segment Summarization.} \label{subsec:seg_summ}
Products may have a large number of reviews.  In our dataset, the single most-reviewed product has a massive \(10,222\) reviews, generating \(90,314\) segments -- completely infeasible to manually process.  Also, many reviews may express the same meaning.
These two characteristics strongly motivate the need for a summarization algorithm to extract popular segments.
Our summarization algorithm should satisfy the following requirements: 1) it must be unsupervised, since we do not have reference summaries; 2) it must be highly scalable since reviews per product regularly exceed \(1,000\) inputs; and 3) it should extract frequently-occurring segments.
In the case of reviews, we observe that the popularity of a segment is generally associated with their frequency of repetition.
If several reviewers talk about a specific segment, e.g., \textit{``display is very good''}, in their reviews for a product, it becomes a popular segment.

To satisfy these requirements, we employ the technique of \citet{angelidis-etal-2021-extractive}.
They train an embedding space consisting of latent codes.
Each latent code is a randomly-initialized vector that groups semantically similar segments.
Then, a later part of the algorithm extracts top segments from each code which are considered popular segments.

Our work uses a slightly different approach to determine segments to extract.
While \citet{angelidis-etal-2021-extractive} use a threshold for the total number of words in the desired output summary, our method emphasizes popularity: we select segments for the output summary which are sampled greater than a tunable threshold $t$ times.
With an overly high $t$ (e.g., $t=50$), too few segments are selected; but if set too low (e.g., $t=5$), the resultant segment quality is often poor and also often syntactically invalid, semantically incomplete or repetitive.
We set the threshold to $t=18$, based on appropriate empirical tuning on our validation set.
For each product, we perform summarization on positive and negative opinions separately.
Table~\ref{table:pos_neg_summary_example} illustrates a few examples of extracted summaries.
\begin{table}\small
    \centering
    \begin{tabularx}{\columnwidth}{|X|X|}
        \hline
        \multicolumn{1}{|c|}{Positive}    &  \multicolumn{1}{|c|}{Negative}    \\
        \hline
        it is great &  battery life is lackluster\\
        the display is awesome  &  camera is not good\\
        screen is great &  does have some issues with clearing memory\\
        meets all expectations  &  it 's just annoying\\
        this tablet is fantastic    &  eventually it refuses to turn on at all\\
        \hline
    \end{tabularx}
    \caption{Extracted Summaries. This table depicts five segments each of positive and negative opinions from a extracted summary of a product.}
    \label{table:pos_neg_summary_example}
\end{table}

\subsection{Rule-based Response Generation}
\label{subsec:templates}
After extracting popular positive and negative segments separately, our final step is to generate the contrastive snippets, conforming to the format exemplified by 
Table~\ref{table:response_example}.

We first analyzed sampled extracted reviews to understand how users actually combine two contrasting opinions when writing their own reviews.
As a result, we inventoried seven common templates that users employ to combine both positive (POS) and negative (NEG) opinions:
\begin{enumerate}
    \item \{POS\} . but , \{NEG\} .
    \item \{POS\} . however \{NEG\} .
    \item \{POS\} . on the other hand , \{NEG\} .
    \item although \{POS\} , according to a few users \{NEG\} .
    \item \{POS\} . yet , some users have also mentioned that \{NEG\} .
    \item \{POS\} . however , there are people who have complained that \{NEG\} .
    \item \{POS\} . on the other hand , a few users have complained that \{NEG\} .
\end{enumerate}
For a given product \(e\), let \(O_{p}\) and \(O_{n}\) represent a set of positive and negative opinions, respectively, extracted by our opinion extraction method.
We combine \(o_{p} \in O_{p}\), and \(o_{n} \in O_{n}\) using the templates, as illustrated in Table~\ref{table:response_example}, to generate an output response.
Then, for each product, we combine each positive segment, \(o_{p}\), with each negative segment, \(o_{n}\), present in the respective extracted summaries.
\begin{table*} \small
    \begin{tabularx}{\linewidth}{|c|X|}
    \hline
    \multirow{2}{*}{Input}  &   \textbf{Positive Opinion}: it works great.\\
        &   \textbf{Negative Opinion}: camera is not good.\\
    \hline
    Response Format &   \{POS\} . However , some users have also mentioned that
\{NEG\} .\\
    \hline
    Generated Response  &   it works great . However , some users have also mentioned that camera is not good .\\
    \hline
    \end{tabularx}
    \caption{Response generation example: a tuple of consisting of a template-based response, generated from an extracted positive and a negative opinion of a product.}
    \label{table:response_example}
\end{table*}

\section{Model Architecture} \label{sec:model_arch}
\begin{figure}
    \centering
    \includegraphics[width=\columnwidth]{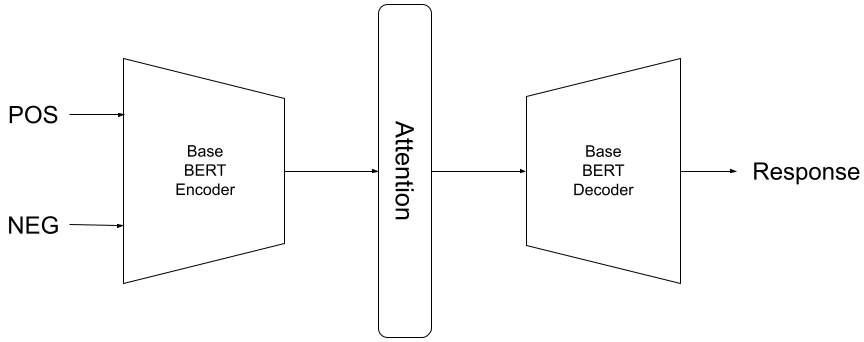}%
    \caption{Model Architecture from \citet{rothe-etal-2020-leveraging}.}
    \label{fig:model_architecture}
\end{figure}

Given a positive opinion \(o_{p} \in O_{p}\), and a negative opinion \(o_{n} \in O_{n}\), our task is to generate a response \(R\) as shown in Table~\ref{table:response_example}.
We use an Encoder--Decoder based architecture similar to \citet{rothe-etal-2020-leveraging}, as depicted in Fig.~\ref{fig:model_architecture}.
For the encoder, we inherit BERT Transformer layer implementations which differs slightly than the canonical Transformer layer implementation \citet{Vaswani2017AttentionIA}; as BERT replaces the standard RELU with GELU activation \cite{Hendrycks2016BridgingNA}.
The implementation of our decoder is also similar to BERT with two modifications:
% Min: why only the left context? why attention?  Justify your decisions.
% Saurabh: Done.
First, the self-attention mechanism is modified to look only at the left context due to the unavailability of right context at the generation time.
Second, an encoder--decoder attention mechanism is added as suggested by \citet{rothe-etal-2020-leveraging}.
We initialize both the encoder and decoder with publicly available pre-trained checkpoints from the uncased base model of BERT to learn and decode hidden representations.
We join \(o_{p}\) and \(o_{n}\), respectively, with a \emph{full stop (.)} to make an input sequence.
We use \emph{mean cross entropy (MCE)} to compute loss.
We fine-tune our model to generate a response fusing a positive and a negative opinion.
In the next section, we describe our experimental settings and analyze results in detail.

\section{Evaluation and Results}
\noindent \textbf{Dataset.}
We use reviews of products from the ``Electronics'' category of Amazon Reviews Dataset (2018) \citet{ni-etal-2019-justifying}.
We generate our dataset in two phases, following the steps in \S~\ref{sec:dataset_generation}.
In the first phase, we consider reviews of \(74\) products only and hand-curate segments in the generated summaries.
We consider syntactically and semantically valid segments only.
In the second phase, we scale the number of products and consider reviews of \(3,269\) products.
After extracting segments from these reviews, we consider only those segments for the summary generation which have part-of-speech patterns similar to hand-curated segments extracted in the first phase.
Thus, we can ensure syntactic validity of the segments.
Our final dataset contains \(174,394\) training instances generated from reviews of \(2,569\) products, \(19,725\) validation instances generated from reviews of \(321\) products, and \(21,397\) test instances generated from reviews of \(379\) products.

\subsection{Implementation Details}
We use the Transformer architecture in segment summarization model and uncased base BERT architecture for our response generation model. \\
% We run all experiments on an NVidia Titan RTX GPU.\\
\\
\noindent \textbf{Segment Summarization.}
As our summarization model is similar to \citet{angelidis-etal-2021-extractive}, we retain their settings in our experiments.
We use a unigram LM SentencePiece vocabulary of size \(32K\)\footnote{\url{https://github.com/google/sentencepiece}} to encode opinion segments.
Our Transformer has a dimension size of \(312\), while its feed-forward layers are of size \(512\).
It uses \(3\) layers and \(4\) internal heads.
The input embedding layer is shared between the encoder and decoder, and \(H = 8\) sentence heads are used to represent every sentence.
For the quantizer, we set number of latent codes \(k = 1024\) and sample \(m = 30\) codes for each segment.
We use the Adam optimizer \cite{Kingma2015AdamAM} with an initial learning rate of \(10^{-3}\) and a learning rate decay of \(0.9\).
We disable segments assignments to latent codes for the first \(4\) epochs as warm--up steps for the Transformer.
We train the model for a total of \(20\) epochs.
At prediction time, in two-step sampling, we sample \(300\) latent codes, and for each code, we sample \(n = 30\) segments.\\
\begin{table*}[t] \small
    \centering
    \begin{tabularx}{\linewidth}{|c|c|c|X|}
    \hline
    Recall  &   \shortstack{Model\\Based}  & \shortstack{Rule\\Based} & \multicolumn{1}{|c|}{Comparison Source}\\
    \hline
    ROUGE-L &   0.9876  &   1.0 &   Input and predicted output\\
    \hline
    \hline
    ROUGE-3 &   0.8563  &   1.0 &   Prediction output and the most similar reference\\
    \hline
    ROUGE-4 &   0.7885  &   1.0 &   Prediction output and the most similar reference\\
    \hline
    \hline
    ROUGE-2 &   0.8376  &   1.0 &   Connecting strings from the prediction output and the most similar reference\\
    \hline
    ROUGE-3 &   0.7884  &   1.0 &   Connecting strings from the prediction output and the most similar reference\\
    \hline
    \end{tabularx}
    \caption{Comparative Snippet Generation Model Evaluation (Column~2; ``Model Based''). The first row (ROUGE-L) measures input information preservation. The next two rows (ROUGE-3 and -4) measure the quality of predicted outputs, and the last two rows' entries measure the quality of the model-proposed connecting strings.}
    \label{tab:metrics}
\end{table*}
\\
\noindent \textbf{Response Generation.}
%
% Min: this is a bit repetitive and contradictory with the previous leading paragraph in S5.1. You say you use it for both the encoder and the decoder but your segment summarization says it uses the transform.
% Saurabh: For summary generation, transformer is used (i have just followed authors' work here) . For comparative text generation, BERT has been used.
Due to the effectiveness of BERT over Transformers in text generation tasks \cite{devlin-etal-2019-bert}, we use the base BERT model for our encoder and decoder.
Since we initialize both the encoder and decoder with uncased base BERT pre-trained checkpoints, our experimental settings are similar to what were used while training the base BERT model.
It has \(12\) layers, hidden size of \(768\), \(12\) attention heads, and vocabulary of \(\sim{30K}\) word pieces.
We fine-tune this model for \(5\) epochs with a batch size of \(32\).
Inputs and outputs are padded to a length of the largest available instance present in training, validation, and test sets.

\subsection{Metrics} \label{subsec:metrics}
An output response should be evaluated on the basis of three aspects:\\
\\
\noindent \textbf{1. Preservation of input information.}
There should not be any change in positive and negative opinions.
The semantic meaning and the syntactic structure of these opinions should be preserved.
We use ROUGE-L to evaluate this aspect.
It measures the longest common subsequence between an output sentence and a reference sentence.
Since we do not modify the positive and negative opinions, the longest common subsequence is identical to one of the input opinions.
For example, for an input sentence \textit{``display is awesome. battery takes long time to charge.''} and the corresponding output sentence \textit{``display is awesome. however, battery takes long time to charge.''}, the longest common sub-sequence is \textit{``display is awesome. battery takes long time to charge.''}\\
\\
\noindent \textbf{2. Quality of output.}
In this aspect, we measure whether the order of words is correct, and connecting string is inserted at the right place.
We use ROUGE-3 and ROUGE-4 to measure this aspect.
These metrics measure the number of common trigrams and quadgrams between a generated output and a reference.
We compare a generated output with each reference separately, and consider the score corresponding to the closest matching reference.\\
\\
\noindent \textbf{3. Quality of connecting string.}
This aspect measures whether words in connecting string are in the correct order and represent a valid sentence connector.
For example, the connecting string \textit{``on the other hand, some users have also mentioned that''} is of higher quality than the string \textit{``on, some users have also mentioned that''}.
We use ROUGE-2 to measure this aspect.
Since a valid connecting string may comprise of part of two or more connecting strings, we do not use ROUGE-3 and ROUGE-4 to avoid heavy penalties.
We remove those tokens from an output that are also present in the input sentence.
We assume that thus remaining sub-string comprises tokens only from connecting strings.
We repeat the same for all the references.
Then we compute the ROUGE-2 metric between the processed output and references.
\subsection{Results and Analysis}
\noindent \textbf{Overall Performance.}
We compute ROUGE-L-recall between input and a generated output to evaluate the model's performance in preserving input information.
As shown in Table~\ref{tab:metrics}, recall of model-based generations is high: 0.9876; yet less than the perfect rule-based generation method that created the dataset.
We consider recall values of ROUGE-3 and ROUGE-4 metrics to measure the quality of generated outputs with respect to the reference outputs. As shown in Table~\ref{tab:metrics}, recall values of ROUGE-3, and ROUGE-4 metrics for model-based generations are 0.8563, and 0.7885, respectively, which we believe is adequate.
Finally, we compute the ROUGE-2 metric specifically confined specifically to the connecting string, in the manner described in ({\it cf.} \S~\ref{subsec:metrics}).
As shown in Table~\ref{tab:metrics}, the recall of ROUGE-2 for our model-based generation is 0.8376.  We believe that the connecting string quality is adequate but can definitely be improved with more careful modelling, and that the connecting string realization is a key component also contributing to overall quality in our second, overall output quality evaluation.
\\
\noindent \textbf{Study on Connective Prediction.}
We examine the connective prediction in more detail as this is the key aspect that is variable in the generation task.  The model's prediction outputs exactly match with one of the references in \(13.09\%\) test cases.
Table~\ref{tab:exact_match} shows distribution of connecting strings corresponding to these exact matches.
Our model not only learns to generate comparative responses by fusing positive and negative opinions but also learns to generate new connecting strings by fusing words of two separate connecting strings or by appending a punctuation symbol with a connecting string.
\begin{table}[!t] \small
    \begin{tabularx}{\columnwidth}{|X|r|}
    \hline
    \centering{Connecting String}  &   \%age\\
    \hline
    but,    &   0.64\%\\
    \hline
    however &   44.68\%\\ 
    \hline
    on the other hand,  &   33.65\%\\
    \hline
    yet, some users have also mentioned that    &   0.86\%\\
    \hline
    although [positive opinion], according to a few users   &   0.57\%\\
    \hline
    on the other hand, a few users have complained that &   10.79\%\\
    \hline
    however, there are people who have complained that  &   8.82\%\\
    \hline
    \end{tabularx}
    \caption{Distribution of connecting strings for which a prediction output matches with one of the references. This distribution is with respect to the total number of exact matches.}
    \label{tab:exact_match}
\end{table}
%
%    \bigskip
%
\begin{table}[] \small
    \begin{tabularx}{\columnwidth}{|X|c|}
    \hline
    \centering{New Connecting String}   & Percent.\\
    \hline
    yet, there are people who have complained that  &   0.07\%\\
    \hline
    but, there are people who have complained that  &   0.07\%\\
    \hline
    however, some users have also mentioned that    &   48.74\%\\
    \hline
    but, some users have also mentioned that        &   0.23\%\\
    \hline
    however,                                        &   48.34\%\\
    \hline
    yet,                                            &   1.51\%\\
    \hline
    but                                             &   0.10\%\\
    \hline
    yet                                             &   0.94\%\\
    \hline
    \end{tabularx}
    \caption{Distribution of new connecting strings. This distribution is with respect to the total number of newly-generated connecting strings.}
    \label{tab:new_conn_string}
\end{table}
%
    %\bigskip
%
\begin{table}[!t] \small
    \begin{tabularx}{\columnwidth}{|X|c|}
    \hline
    \multicolumn{1}{|c|}{Error Type} &   Percent.\\
    \hline
    Incorrect mixing    &   48.22\%\\
    \hline
    Missing although word   &   17.82\%\\
    \hline
    Single word ``on'' insertion  &   14.65\%\\
    \hline
    Input information modification  &   19.26\%\\
    \hline
    \end{tabularx}
    \caption{Common generation errors from our pre-trained BERT model's output.}
    \label{tab:top_error_types}
\end{table}
%\vfill
%
Our model fuses an additional \(13.94\%\) test cases with newly-generated connecting strings.
Table~\ref{tab:new_conn_string} shows the distributional analysis of such newly-generated connecting strings.
We can see that \textit{``however, some users have also mentioned that''} occurs the maximum number of times in such test cases and has been generated by fusing \textit{``however''}, and \textit{``some users have also mentioned that''}.
The second most-frequent generated string is \textit{``however,''} in which a comma (\textit{``,''}) has been appended to a connecting string.
\begin{table*}[h] \small
    \begin{tabularx}{\linewidth}{|X|c|}
    \hline
    \multicolumn{1}{|c|}{Incorrect Mixing Pattern}  &   Distribution\\
    \hline
    however the other hand, [first word from the negative opinion] few users have complained that &   9.85\%\\
    \hline
    however [first word from the negative opinion] there are people who have complained that &   7.83\%\\
    \hline
    on the other hand, [first word from the negative opinion] few users have complained that &   5.83\%\\
    \hline
    on [first word from the negative opinion] there are people who have complained that    &   7.38\%\\
    \hline
    on, some users have also mentioned that &   8.15\%\\
    \hline
    however the other hand, &   6.40\%\\
    \hline
    \end{tabularx}
    \caption{Top incorrect mixing patterns. Here percentage is w.r.t. all the failure cases.}
    \label{tab:top_incorrect_mixing_pattern}
    \bigskip
    \begin{tabularx}{\textwidth}{|c|c|X|}
    \hline
    \multirow{2}{*}{\shortstack{Incorrect\\mixing}}   &  Expected &   the entire set is comfortable. on the other hand, a few users have complained that right side slides down.\\
    \cline{2-3}
        &   Predicted   & the entire set is comfortable. on the other hand, right few users have complained that right side slides down.\\
    \hline
    \multirow{2}{*}{\shortstack{Missing\\``although''}}   &  Expected &   although the retractil system works fine, according to a few users the pads are sort of squarish.\\
    \cline{2-3}
        &   Predicted   & the the retractil system works fine, according to a few users the pads are sort of squarish.\\
    \hline
    \multirow{2}{*}{\shortstack{Insertion\\of ``on''}}   &  Expected &   the 415's are a great upgrade from the oem earbuds. but, it is super uncomfortable.\\
    \cline{2-3}
        &   Predicted   & the 415's are a great upgrade from the oem earbuds. on, it is super uncomfortable.\\
    \hline
    \multirow{2}{*}{\shortstack{Information\\modification}}   &  Expected &   sound is pretty good. but, the movement is actually more like a saw.\\
    \cline{2-3}
        &   Predicted   & sound is pretty good. however, the movement is actually more like a see.\\
    \hline
    \end{tabularx}
    \caption{Examples of top errors.}
    \label{tab:top_error_examples}
\end{table*}

Our model also generates incorrectly fused sentences.
Table~\ref{tab:top_error_types} shows main types of error.
The maximum number of failure cases occur due to incorrect mixing of parts of different connecting strings.
In such cases, either an extra word is inserted, or more words are missing from the connecting string.
Table~\ref{tab:top_incorrect_mixing_pattern} depicts top incorrect mixing patterns.
In the four patterns, we can see that the first word of negative opinion is inserted in between the connecting string.
Our analysis shows that it happens due to the ambiguity present in our training dataset.
For example, our training dataset contains connecting strings \textit{``however''}, and \textit{``however, there are people who have complained that''}.
In the case of the former, just after the connecting string \textit{``however''}, the model inserts words from the negative opinion.
While in the latter case, words from a connecting string are inserted after \text{``however''}.
Therefore, we assume that at prediction time probability of inserting the first word of the negative opinion after \textit{``however''} becomes highest, thus resulting in an incorrect mixing of connecting strings.
A similar argument exists for the incorrect cases containing string \textit{``on the other hand''}.
Table~\ref{tab:top_error_examples} shows an example of incorrectly mixed connecting string.

The second most common failure case is associated with input information modification.
Ideally, a comparative output sentence must contain positive and negative opinions without modification.
But our model, sometimes, generates an output sentence that either deletes or repeats one or more words from the input or replaces a word with its synonym or base form (Table~\ref{tab:top_error_examples}).
As mentioned in the section ({\it cf.} \S~\ref{subsec:templates}), except one, in all other templates connecting string is inserted between the positive and negative opinions.
In case of \textit{although {POS}, according to a few users {NEG}}, we also prepend a word \textit{``although''} in the output sentence.
Since most of the training instances insert a connecting string only in between, our model does not learn properly to prepend a word \textit{``although''} and thus gives rise to the third most occurring failure cases in which the first word of the positive opinion is repeated instead of prepending a word \textit{``although''}.
An example of such a case has been shown in Table~\ref{tab:top_error_examples}.
The last most occurring errors are associated with single word insertion \textit{``on''} between the positive and negative opinions, as shown in Table~\ref{tab:top_error_examples}.\\

\noindent \textbf{Why not use rules to generate responses if these give better performance?}
As shown in our results, rule-based generations outperform model-based generations.
Therefore, an obvious question arises on the use of model-based generations.
Templates used for generating rule-based generations have been manually selected from a random analysis of reviews.
But, in the future, we want our model to automatically learn styles of comparative response generations from the given dataset and use these styles to fuse positive and negative opinions.
Therefore, we prefer to use model-based generations and improve their accuracy.
\section{Conclusion and Future Work}
We introduced a novel task of generating a comparative response (or ``snippet'') regarding a product that combines positive and negative opinions together in a single sentence.
As such comparative responses are not easily found in natural review environments, we generate such comparative responses through extractive summaries of product reviews using an unsupervised approach.  To spur future research in this area, we have also made our dataset public and leveraged the prior Amazon reviews corpus, popular with the research community. 
Throughout our work we assume that all reviews are genuine and have been written by buyers who have used the product.
We investigate and benchmark a baseline model for this task that combines state-of-the-art text representation (BERT) in an encoder--decoder architecture to generate a comparative response.
Our analysis of the output results shows that even such a state-of-the-art pre-trained model does not generate perfect responses.

There are limitations of our work that we hope to address in the future.
Currently, positive and negative opinions in a generated response may or may not be related to the same aspect.
As such, improvements to better generate more naturalistic responses may restrict generation to opinions where both the positive and negative discuss the same product aspect.
% Min: this is more like a limitation that can be addressed later.  Consider moving to the conclusion.
% Saurabh: Done.
%Secondly, throughout our work we assume that all reviews are genuine and have been written by buyers who have used the product. This is unlikely to be true in real-world scenarios, 
In future we would also like to quantify the veracity of the opinions and weight them accordingly.

\section*{Acknowledgements}
We acknowledge the support of NVIDIA Corporation for their donation of the Titan X GPU that facilitated this research.

\bibliography{anthology,custom}

\begin{thebibliography}{20}
\expandafter\ifx\csname natexlab\endcsname\relax\def\natexlab#1{#1}\fi

\bibitem[{Amplayo and Lapata(2021)}]{Amplayo2021InformativeAC}
Reinald~Kim Amplayo and Mirella Lapata. 2021.
\newblock Informative and controllable opinion summarization.
\newblock In \emph{EACL}.

\bibitem[{Angelidis et~al.(2021)Angelidis, Amplayo, Suhara, Wang, and
  Lapata}]{angelidis-etal-2021-extractive}
Stefanos Angelidis, Reinald~Kim Amplayo, Yoshihiko Suhara, Xiaolan Wang, and
  Mirella Lapata. 2021.
\newblock \href {https://doi.org/10.1162/tacl_a_00366} {Extractive opinion
  summarization in quantized transformer spaces}.
\newblock \emph{Transactions of the Association for Computational Linguistics},
  9:277--293.

\bibitem[{Angelidis and Lapata(2018)}]{angelidis-lapata-2018-multiple}
Stefanos Angelidis and Mirella Lapata. 2018.
\newblock \href {https://doi.org/10.1162/tacl_a_00002} {Multiple instance
  learning networks for fine-grained sentiment analysis}.
\newblock \emph{Transactions of the Association for Computational Linguistics},
  6:17--31.

\bibitem[{Ben-David et~al.(2020)Ben-David, Keller, Malmi, Szpektor, and
  Reichart}]{ben-david-etal-2020-semantically}
Eyal Ben-David, Orgad Keller, Eric Malmi, Idan Szpektor, and Roi Reichart.
  2020.
\newblock \href {https://doi.org/10.18653/v1/2020.findings-emnlp.135}
  {Semantically driven sentence fusion: Modeling and evaluation}.
\newblock In \emph{Findings of the Association for Computational Linguistics:
  EMNLP 2020}, pages 1491--1505, Online. Association for Computational
  Linguistics.

\bibitem[{Devlin et~al.(2019)Devlin, Chang, Lee, and
  Toutanova}]{devlin-etal-2019-bert}
Jacob Devlin, Ming-Wei Chang, Kenton Lee, and Kristina Toutanova. 2019.
\newblock \href {https://doi.org/10.18653/v1/N19-1423} {{BERT}: Pre-training of
  deep bidirectional transformers for language understanding}.
\newblock In \emph{Proceedings of the 2019 Conference of the North {A}merican
  Chapter of the Association for Computational Linguistics: Human Language
  Technologies, Volume 1 (Long and Short Papers)}, pages 4171--4186,
  Minneapolis, Minnesota. Association for Computational Linguistics.

\bibitem[{Diao et~al.(2014)Diao, Qiu, Wu, Smola, Jiang, and
  Wang}]{Diao2014JointlyMA}
Qiming Diao, Minghui Qiu, Chao-Yuan Wu, Alex Smola, Jing Jiang, and Chong Wang.
  2014.
\newblock Jointly modeling aspects, ratings and sentiments for movie
  recommendation (jmars).
\newblock \emph{Proceedings of the 20th ACM SIGKDD international conference on
  Knowledge discovery and data mining}.

\bibitem[{Feng and Hirst(2014)}]{feng-hirst-2014-linear}
Vanessa~Wei Feng and Graeme Hirst. 2014.
\newblock \href {https://doi.org/10.3115/v1/P14-1048} {A linear-time bottom-up
  discourse parser with constraints and post-editing}.
\newblock In \emph{Proceedings of the 52nd Annual Meeting of the Association
  for Computational Linguistics (Volume 1: Long Papers)}, pages 511--521,
  Baltimore, Maryland. Association for Computational Linguistics.

\bibitem[{Geva et~al.(2019)Geva, Malmi, Szpektor, and
  Berant}]{geva-etal-2019-discofuse}
Mor Geva, Eric Malmi, Idan Szpektor, and Jonathan Berant. 2019.
\newblock \href {https://doi.org/10.18653/v1/N19-1348} {{D}isco{F}use: A
  large-scale dataset for discourse-based sentence fusion}.
\newblock In \emph{Proceedings of the 2019 Conference of the North {A}merican
  Chapter of the Association for Computational Linguistics: Human Language
  Technologies, Volume 1 (Long and Short Papers)}, pages 3443--3455,
  Minneapolis, Minnesota. Association for Computational Linguistics.

\bibitem[{Gupta et~al.(2019)Gupta, Kulkarni, Chanda, Rayasam, and
  Lipton}]{Gupta2019AmazonQAAR}
Mansi Gupta, Nitish Kulkarni, Raghuveer Chanda, Anirudha Rayasam, and
  Zachary~Chase Lipton. 2019.
\newblock Amazonqa: A review-based question answering task.
\newblock In \emph{IJCAI}.

\bibitem[{Hendrycks and Gimpel(2016)}]{Hendrycks2016BridgingNA}
Dan Hendrycks and Kevin Gimpel. 2016.
\newblock Bridging nonlinearities and stochastic regularizers with gaussian
  error linear units.
\newblock \emph{ArXiv}, abs/1606.08415.

\bibitem[{Kim(2014)}]{Kim2014ConvolutionalNN}
Yoon Kim. 2014.
\newblock Convolutional neural networks for sentence classification.
\newblock In \emph{EMNLP}.

\bibitem[{Kingma and Ba(2015)}]{Kingma2015AdamAM}
Diederik~P. Kingma and Jimmy Ba. 2015.
\newblock Adam: A method for stochastic optimization.
\newblock \emph{CoRR}, abs/1412.6980.

\bibitem[{Ni et~al.(2019)Ni, Li, and McAuley}]{ni-etal-2019-justifying}
Jianmo Ni, Jiacheng Li, and Julian McAuley. 2019.
\newblock \href {https://doi.org/10.18653/v1/D19-1018} {Justifying
  recommendations using distantly-labeled reviews and fine-grained aspects}.
\newblock In \emph{Proceedings of the 2019 Conference on Empirical Methods in
  Natural Language Processing and the 9th International Joint Conference on
  Natural Language Processing (EMNLP-IJCNLP)}, pages 188--197, Hong Kong,
  China. Association for Computational Linguistics.

\bibitem[{Rothe et~al.(2020)Rothe, Narayan, and
  Severyn}]{rothe-etal-2020-leveraging}
Sascha Rothe, Shashi Narayan, and Aliaksei Severyn. 2020.
\newblock \href {https://doi.org/10.1162/tacl_a_00313} {Leveraging pre-trained
  checkpoints for sequence generation tasks}.
\newblock \emph{Transactions of the Association for Computational Linguistics},
  8:264--280.

\bibitem[{Socher et~al.(2011)Socher, Pennington, Huang, Ng, and
  Manning}]{socher-etal-2011-semi}
Richard Socher, Jeffrey Pennington, Eric~H. Huang, Andrew~Y. Ng, and
  Christopher~D. Manning. 2011.
\newblock \href {https://aclanthology.org/D11-1014} {Semi-supervised recursive
  autoencoders for predicting sentiment distributions}.
\newblock In \emph{Proceedings of the 2011 Conference on Empirical Methods in
  Natural Language Processing}, pages 151--161, Edinburgh, Scotland, UK.
  Association for Computational Linguistics.

\bibitem[{Suhara et~al.(2020)Suhara, Wang, Angelidis, and
  Tan}]{suhara-etal-2020-opiniondigest}
Yoshihiko Suhara, Xiaolan Wang, Stefanos Angelidis, and Wang-Chiew Tan. 2020.
\newblock \href {https://doi.org/10.18653/v1/2020.acl-main.513}
  {{O}pinion{D}igest: A simple framework for opinion summarization}.
\newblock In \emph{Proceedings of the 58th Annual Meeting of the Association
  for Computational Linguistics}, pages 5789--5798, Online. Association for
  Computational Linguistics.

\bibitem[{Tang et~al.(2015)Tang, Qin, and Liu}]{tang-etal-2015-document}
Duyu Tang, Bing Qin, and Ting Liu. 2015.
\newblock \href {https://doi.org/10.18653/v1/D15-1167} {Document modeling with
  gated recurrent neural network for sentiment classification}.
\newblock In \emph{Proceedings of the 2015 Conference on Empirical Methods in
  Natural Language Processing}, pages 1422--1432, Lisbon, Portugal. Association
  for Computational Linguistics.

\bibitem[{Turney(2002)}]{turney-2002-thumbs}
Peter Turney. 2002.
\newblock \href {https://doi.org/10.3115/1073083.1073153} {Thumbs up or thumbs
  down? semantic orientation applied to unsupervised classification of
  reviews}.
\newblock In \emph{Proceedings of the 40th Annual Meeting of the Association
  for Computational Linguistics}, pages 417--424, Philadelphia, Pennsylvania,
  USA. Association for Computational Linguistics.

\bibitem[{van~den Oord et~al.(2017)van~den Oord, Vinyals, and
  Kavukcuoglu}]{Oord2017NeuralDR}
A{\"a}ron van~den Oord, Oriol Vinyals, and Koray Kavukcuoglu. 2017.
\newblock Neural discrete representation learning.
\newblock In \emph{NIPS}.

\bibitem[{Vaswani et~al.(2017)Vaswani, Shazeer, Parmar, Uszkoreit, Jones,
  Gomez, Kaiser, and Polosukhin}]{Vaswani2017AttentionIA}
Ashish Vaswani, Noam~M. Shazeer, Niki Parmar, Jakob Uszkoreit, Llion Jones,
  Aidan~N. Gomez, Lukasz Kaiser, and Illia Polosukhin. 2017.
\newblock Attention is all you need.
\newblock \emph{ArXiv}, abs/1706.03762.

\end{thebibliography}
\bibliographystyle{acl_natbib}

% \appendix

% \section{Example Appendix}
% \label{sec:appendix}

% This is an appendix.

\end{document}